\pdfoutput=1

\documentclass[11pt]{article}

\usepackage[]{EMNLP2022}

\usepackage{times}
\usepackage{latexsym}

\usepackage[T1]{fontenc}

\usepackage[utf8]{inputenc}

\usepackage{microtype}
\usepackage{inconsolata}

\usepackage[toc,page]{appendix}
\usepackage{colortbl}

\hypersetup{
	linkcolor=magenta,
	filecolor=magenta,
}

\usepackage{times}
\usepackage{latexsym}
\usepackage{graphicx}
\usepackage{float}
\usepackage{amsmath}
\usepackage{amsfonts}
\usepackage{subfigure}
\usepackage[ruled,vlined]{algorithm2e}

\usepackage{booktabs}
\usepackage{placeins}
\usepackage{makecell}
\usepackage{bm}
\usepackage{multirow}
\usepackage{multicol}
\usepackage{arydshln}
\usepackage{url}
\usepackage{tablefootnote}
\usepackage[normalem]{ulem}

\usepackage{amssymb}
\usepackage{pifont}
\definecolor{mycolor1}{HTML}{495464}
\definecolor{mycolor2}{HTML}{bbbfca}
\definecolor{mycolor3}{HTML}{e8e8e8}
\definecolor{mycolor4}{HTML}{f4f4f2}

\newcommand{\tabref}[1]{Table~\ref{#1}}

\usepackage{CJKutf8} 

\title{Exploring the Usage of Chinese Pinyin in Pretraining}

 \author{Baojun Wang, Kun Xu, Lifeng Shang \\
	\{puking.w, nukux.xu, Shang.Lifeng\}@huawei.com}

\begin{document}
\begin{CJK*}{UTF8}{gbsn}
\maketitle

\begin{abstract}
Unlike alphabetic languages, Chinese spelling and pronunciation are different. Both characters and pinyin take an important role in Chinese language understanding. 
In Chinese NLP tasks, we almost adopt characters or words as model input, and few works study how to use pinyin. 
However, pinyin is essential in many scenarios, such as error correction and fault tolerance for ASR-introduced errors.
Most of these errors are caused by the same or similar pronunciation words, and we refer to this type of error as SSP(the same or similar pronunciation) errors for short.
In this work, we explore various ways of using pinyin in pretraining models and propose a new pretraining method called PmBERT.
Our method uses characters and pinyin in parallel for pretraining. Through delicate pretraining tasks, the characters and pinyin representation are fused, which can enhance the error tolerance for SSP errors.
We do comprehensive experiments and ablation tests to explore what makes a robust phonetic enhanced Chinese language model. The experimental results on both the constructed noise-added dataset and the public error-correction dataset demonstrate that our model is more robust compared to SOTA models.
\end{abstract}

\section{Introduction}

Benefits from the large-scale pretraining, language models have dominated in many Chinese NLP tasks \cite{brown2020language,reimers2019sentence,zhang2019bertscore, liu2019text, sun2019fine}.
Compared to character, pinyin has drawn less attention in recently Chinese pretrain language models. 
However, pinyin conveys more pronunciation information and it is essential in many scenarios such as dealing with SSP errors.

\begin{figure}
    \centering
    \scalebox{0.45}{
    \includegraphics{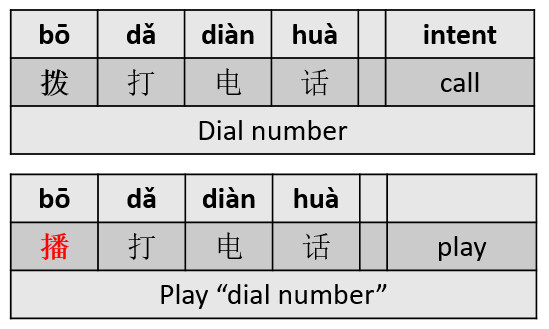}
    }
    \caption{In Chinese, even change a signle token to a phonetic similar token will alter the meaning of sentence completely.}
    \label{fig:my_label}
\end{figure}

Chinese has roughly 400 pinyin and 3500 commonly used characters, most of which share similar pronunciation. Different from alphabetic languages, it cannot be entered into systems without the help of Automatic Speech Recognition(ASR) or pinyin. These methods often suffer from phonetic errors, most of which are SSP errors. As shown in Figure 1, character "拨" and character "播" have the same pronunciation "bo". User intent will be changed from "make a call" to "play a media file" if the ASR system misrecognizes  "拨" to  "播". Robustness to these errors is important in realistic applications.

To increase the robustness, a few works have made their attempts to incorperate pinyin into pretraining models to bridge the relation between semantics and phonetics: changing pretraining mask strategies \cite{liu-etal-2021-plome, zhang2021correcting, sun2021chinesebert}, using extra pinyin encoders \cite{huang2021phmospell, xu2021read}. These methods show impressive performance on specific tasks such as Chinese spell correction. To the best of our knowledge, there are no existing works that fairly compare their capability in different downstream tasks. Especially there lack of public datasets to measure the model robustness to the SSP problem. How to efficiently utilize pinyin in Chinese language models is still an open question.

In this paper, we explore various ways of using pinyin with pretraining models, create a robustness test dataset, and conduct adequate experiments to compare these methods. Meanwhile, we propose a new pretraining method with pinyin called PmBERT(Pinyin mixed BERT). 
In summary, our contributions are as follows:

\begin{itemize}
    \item [1)] 
    We fairly compare current methods which incorperate pinyin into Chinese language model under different NLP tasks. Comprehensive explorations and comparisons of phonetic information are presented.
    \vspace{-0.2cm}
    \item [2)] 
    We propose a new pretraining model called PmBERT to jointly learn semantic and phonetic representations. Extensive experiments were done to verify the robustness of our proposed model.
    \vspace{-0.2cm}
    \item [3)]
    We create a robustness test dataset to measure SOTA model performance. Our pretraining model and the test dataset will public soon to drive further research.
\end{itemize}

\section{Why Pinyin is Used Less}
In Chinese natural language processing, the basic models use Chinese characters or words as their input. Pinyin is rarely adopted in the era of pretraining, except for some special tasks(e.g. Chinese input method).
We summarize main reasons as follows:

\paragraph{Pinyin sequence is longer.}
Since pinyin consists of three parts(initials, finals, and tones), its sequence is generally much longer than the character sequence. The longer the sequence is, the longer compute time models take. When using pretraining models, we suffers more computing burden due to the $O(n^2)$ time complexity attention mechanism.


\paragraph{Pinyin is lack of readability.}
For example, in a pinyin sequence shown in Figure 1, "zhōng wén shì yī mén shí fèn bó dà jīng shēn de yǔ yán", it is difficult to understand the meaning of the whole sentence at a glance. Many tasks, such as named entity extraction, cannot be used directly if the pinyin is output, and still have to be converted to text, which is not very acceptable in reality.

Although pinyin is not as convenient as characters, it is indirectly involved in many industrial scenarios. Take ASR for example. It aims to recognize speech into text, but this process is extremely susceptible to environmental noise and is prone to misrecognition. Once the word is misrecognized, it will directly affect the downstream tasks. For example, in intelligent assistants, this error would lead to incorrect intent recognition. Another important scenario is query. Chinese queries are generally input through the pinyin input method, which would contain SSP errors. If the pinyin information can be introduced into the NLP model, the above mentioned homophone and near-sound errors can be alleviated.

\begin{figure}
    \centering
    \includegraphics[width=7.5cm]{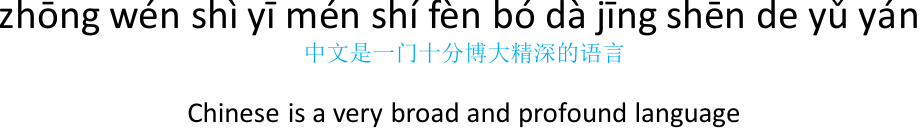}
    \caption{Pinyin sentences are not intuitive, and it is difficult to know what they mean at a glance.}
    \label{fig:readablity}
\end{figure}

\section{Approach}
In this section, we introduce our PmBERT model structure and pretraining task.
We construct our model based on BERT \citet{devlin2018bert} architecture.
We design a new masking strategy to bridge the relation between the phonetic and semantic features.

\subsection{Model Architectures}
BERT has shown its impressive performance in various NLP tasks.
Currently, it is one of the best context-aware semantic encoders.
In this work, we choose BERT as our  PmBERT backbone.
We modify the embeddings layer to utilize pinyin features.
With this minor change, we fuse BERT semantic representation with pinyin phonetic representations.
\begin{figure}[]
    \centering
    \includegraphics[width=0.4\textwidth,height=10cm]{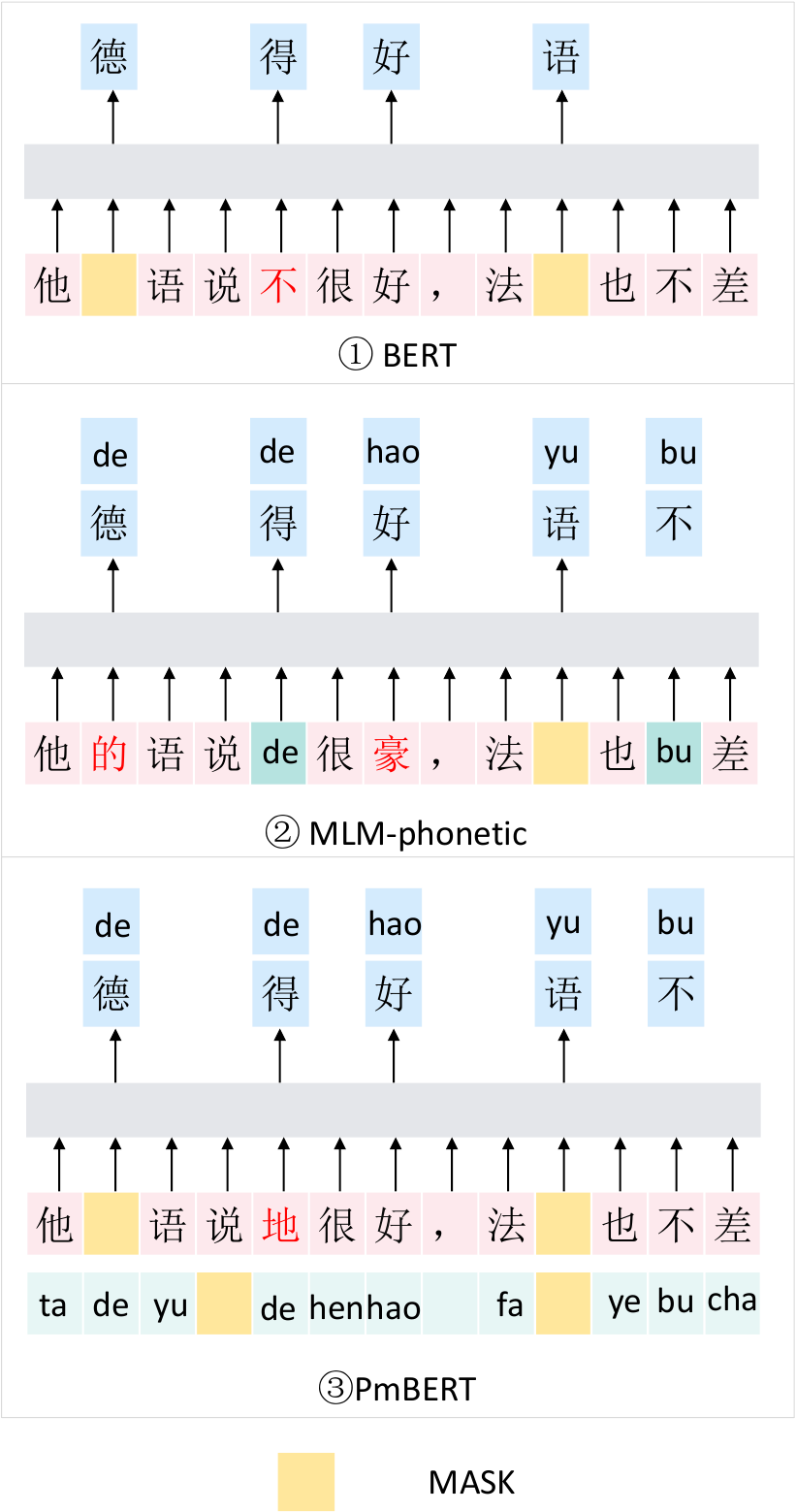}
    \caption{A demonstration of PmBERT masking strategy compared with BERT and MLP-phonetic. PmBERT adopt pinyin in parallel.}
    \label{mixed-or-parallel}
\end{figure}

\paragraph{Embeddings}
Similar to BERT, we use wordpiece embeddings with a 20000+ token vocabulary to represent character-level features.
We convert each Chinese character in the vocab to pinyin with a dictionary and create a pinyin embeddings to pretrain their phonetic level features.
For those non-Chinese tokens, we use a special token([UNK]) to hold the place. To make our method can be simply used in pretraining models, we do not use any extra encoder to fuse semantic and phonetic features. 
Similar to the construction of BERT embeddings, we add character embeddings, pinyin embeddings, position embeddings, and segment embeddings together to represent a given token.

\paragraph{Encoders}
PmBERT's encoder is a multi-layer bidirectional Transformer encoder just the same as BERT's encoder. Specifically, we use choose BERT$_{base}$ structure as our backbone which is a 12-layer transformer with 12 heads and 768 hidden dimensions.

\subsection{Masking scheme}
BERT replaces 15\% tokens with a special token([MASK]) at random and tries to reconstruct the original input to learn semantic features. We expand this idea to pinyin space. We construct a pinyin confusion set first. Then we mask part of Chinese characters with a special token or replace them with their confusion characters. A cross entropy loss is applied to recover the original tokens. The example of our mask scheme is illustrated in Figure \ref{mixed-or-parallel}.

\paragraph{Pretrain pinyin confusion set} 
Unlike the conventional method to construct confusion set with human experience \cite{liu-etal-2021-plome} to pretrain model, for a given Chinese character we use characters that share the same pinyin as its confusion set. Firstly, we calculate each Chinese character's usage frequency through our corpus. Then we delete those Chinese characters which do not exist in the original BERT vocab and normalize remain words usage frequency. After that, we get a confusion set with pinyin inputs and their frequency.

\paragraph{Masked LM}
For a given input subword sequence, BERT chooses 15\% of the token position to mask. Specifically, they mask the token with 1) a special token ([MASK]) 80\% of the time 2) a random token 10\% of the time 3) an unchanged token 10\% of the time. 

We expand this idea to pinyin space. In the setting of our model, a Chinese token is represented by a token embedding and a pinyin embedding. To make our models can understand pinyin and character features, we design a task to mask pinyin and tokens together and recover them through their context pinyin and token. To fuse pinyin features and character features, we design task 1) to mask token only and recover token with pinyin 2) mask pinyin only and recover pinyin with token. To make our model robust to phonetic errors, we replace tokens from their confusion set and try to recover them with context and their pinyin. See 
Algorithm \ref{alg:masking_strategy} for detail.

\begin{table*}[]
\centering
\scalebox{0.82}{
\begin{tabular}{lcccccccccc}
\toprule
\multirow{2}{*}{Method} & \multirow{2}{*}{Pretrain} & \multirow{2}{*}{Tone} & \multicolumn{2}{c}{Where} & \multicolumn{2}{c}{Fusion} & \multicolumn{2}{c}{Mask} & \multicolumn{2}{c}{Task} \\ \cmidrule{4-11} 
 &  &  & \multicolumn{1}{c}{input} & out & \multicolumn{1}{c}{concat} & add & \multicolumn{1}{c}{pinyin} & cofset & \multicolumn{1}{c}{special} & general \\ \midrule
PLOME \cite{liu-etal-2021-plome} & \checkmark &  & \multicolumn{1}{c}{\checkmark} &  & \multicolumn{1}{c}{} & \checkmark & \multicolumn{1}{c}{} & \checkmark & \multicolumn{1}{c}{\checkmark} &  \\ \midrule
PHMOSpell \cite{huang2021phmospell} &  & \checkmark & \multicolumn{1}{c}{\checkmark} &  & \multicolumn{1}{c}{} & \checkmark & \multicolumn{1}{c}{} & \checkmark & \multicolumn{1}{c}{\checkmark} &  \\ \midrule
ReadListenAndSee \cite{xu2021read} &  & \checkmark & \multicolumn{1}{c}{\checkmark} &  & \multicolumn{1}{c}{} & \checkmark & \multicolumn{1}{c}{\checkmark} &  & \multicolumn{1}{c}{\checkmark} &  \\ \midrule
MLM-phonetics \cite{zhang2021correcting} & \checkmark &  & \multicolumn{1}{c}{\checkmark} &  & \multicolumn{1}{c}{} & \checkmark & \multicolumn{1}{c}{\checkmark} & \checkmark & \multicolumn{1}{c}{\checkmark} &  \\ \midrule
PERT \cite{pbert} & \checkmark &  & \multicolumn{1}{c}{\checkmark} &  & \multicolumn{1}{c}{} &  & \multicolumn{1}{c}{\checkmark} &  & \multicolumn{1}{c}{} & \checkmark \\ \midrule
Pinyin-fuse-tail$^∗$ &  &  & \multicolumn{1}{c}{} & \checkmark & \multicolumn{1}{c}{\checkmark} &  & \multicolumn{1}{c}{\checkmark} &  & \multicolumn{1}{c}{} & \checkmark \\ \midrule
Pinyin-concat-header$^∗$ &  &  & \multicolumn{1}{c}{\checkmark} &  & \multicolumn{1}{c}{\checkmark} &  & \multicolumn{1}{c}{\checkmark} &  & \multicolumn{1}{c}{} & \checkmark \\ \midrule
Pinyin-add-header$^∗$ &  &  & \multicolumn{1}{c}{\checkmark} &  & \multicolumn{1}{c}{} & \checkmark & \multicolumn{1}{c}{\checkmark} &  & \multicolumn{1}{c}{} & \checkmark \\ \midrule
ChineseBert \cite{sun2021chinesebert} & \checkmark & \checkmark & \multicolumn{1}{c}{\checkmark} &  & \multicolumn{1}{c}{} & \checkmark & \multicolumn{1}{c}{\checkmark} &  & \multicolumn{1}{c}{} & \checkmark \\ \midrule
PmBERT & \checkmark &  & \multicolumn{1}{c}{\checkmark} &  & \multicolumn{1}{c}{} & \checkmark & \multicolumn{1}{c}{\checkmark} & \checkmark & \multicolumn{1}{c}{} & \checkmark \\ \bottomrule
\end{tabular}}

\caption{Comparison among SOTA pinyin enhanced pretrained models and their usage of pinyin.}
\label{use-way}
\end{table*}
\section{Experiments}
\subsection{Overview}
\paragraph{Pretrain setup} We use Chinese Wikipedia and People’s Daily up to 12GB corpus to pretrain our PmBERT. 
Corpus is tokenized to 128 sequence-length sentences.
Since we do not adopt NSP as our pretraining task, we only use sentences from the same document for pretraining.
To avoid using the same mask for each training instance, a dynamic masking strategy is used during the training process. 
We train our PmBERT 0.4 million steps on 16 NVIDIA Tesla V100 GPUs for 4 days with batch size 1024, learning rate 1e-4, 10\% dropout ratio, 10\% warm up ratio, \cite{loshchilov2017decoupled} optimizer and 0.01 decay rate.

\paragraph{Models using pinyin}
There are various ways of using pinyin in Chinese NLP models, and we summarize these methods shown in \tabref{use-way} according to whether to pretrain, use tone or not, fusion ways, mask strategy, and generality. For the masking strategy, part of current pretraining works indirectly uses pinyin. Unlike BERT uses a special token([MASK]), these works construct a confusion set by phonetic characters and use characters from the confusion set to mask. Another part of the works directly uses pinyin to replace the original inputs. 

\subsection{Baselines}
According to the different ways of using pinyin, we choose the strongest method(\citet{liu-etal-2021-plome, zhang2021correcting, xu2021read, pbert, sun2021chinesebert}) as our baseline. 
The ways of using pinyin for all methods refer to Table 2. 
The model\footnote{\url{https://mp.weixin.qq.com/s/JyXN9eukS-5XKvcJORTobg}} with label * is from a technical report which is not public, and it is the result of our own reproduction.
Here, pinyin-fuse-tail fuse pinyin embeddings with token features encoded by BERT. Pinyin-add-header* add token features and pinyin features in the embedding layer. Pinyin-add-header concat token features and pinyin features first then project to 768 hidden in the embedding layer. To compare the enhancement of error correction and error tolerance after the introduction of pinyin, we also use the related adversarial model as our baseline.

\subsection{Datasets}
To evaluate the enhancement effect of the introduction of pinyin on the model, we verify it on the classification task, the sequence labeling task, and Chinese spell correction task respectively. The datasets are divided into two categories, one is the noise-added data set constructed by ourselves, and the other is the Chinese public error correction data set. 

We conduct our experiments on four Chinese datasets, MSRA~\citep{levow-2006-third}, Sina Weibo~\citep{peng-dredze-2015-named}, SMP2017~\citep{zhang2017first} and Toutiao News\footnote{\url{https://github.com/aceimnorstuvwxz/toutiao-text-classfication-dataset}}.
More dataset details refer to Appendix \ref{apd:datasets}.

To build the noisy dataset, we replace part of the Chinese tokens with phonetic similar tokens. Note that, we do not adpot the confusion set used in pretrainging stage to avoid information leak.
To fairly compare models, we build the phonetically similar confusion set from a humman collected 8-million annotated ASR data.
The clean text and transcribed texts are aligned so that we can extract confused words through the data.
We also include a popularly used phonetically similar confusion set\footnote{\url{https://github.com/contr4l/SimilarCharacter}} into ours. This dataset will be public soon.

\begin{table*}[h]
\centering
\scalebox{1.0}{
\begin{tabular}{lcccccc}
\toprule
 & \multicolumn{1}{c}{Weibo} & \multicolumn{1}{c}{MSRA} & \multicolumn{1}{c}{SMP2017} & \multicolumn{1}{c}{Toutiao} &  & Sighan15 \\ \midrule
Model & \multicolumn{4}{c}{20\%noise} & AVG &  \\ \midrule
BERT \cite{devlin2018bert} & \multicolumn{1}{c}{49.50} & \multicolumn{1}{c}{71.61} & \multicolumn{1}{c}{80.57} & \multicolumn{1}{c}{73.28} & 68.74 & 74.30 \\ \midrule
BERT-adv \cite{advbert} & \multicolumn{1}{c}{48.56} & \multicolumn{1}{c}{72.95} & \multicolumn{1}{c}{83.18} & \multicolumn{1}{c}{75.45} & 70.04 & 75.60 \\ \midrule
PERT$^*$ \cite{pbert} & \multicolumn{1}{c}{40.21} & \multicolumn{1}{c}{64.25} & \multicolumn{1}{c}{73.78} & \multicolumn{1}{c}{69.45} & 61.92 & 67.98 \\ \midrule
Pinyin-fuse-tail$^*$ & \multicolumn{1}{c}{44.82} & \multicolumn{1}{c}{75.41} & \multicolumn{1}{c}{83.96} & \multicolumn{1}{c}{79.09} & 70.82 & 74.91 \\ \midrule
Pinyin-concat-header$^*$ & \multicolumn{1}{c}{44.79} & \multicolumn{1}{c}{83.92} & \multicolumn{1}{c}{87.24} & \multicolumn{1}{c}{80.62} & 74.14 & 76.12 \\ \midrule
Pinyin-add-header$^*$ & \multicolumn{1}{c}{48.64} & \multicolumn{1}{c}{69.85} & \multicolumn{1}{c}{87.03} & \multicolumn{1}{c}{78.67} & 71.05 & 75.64 \\ \midrule
ChineseBERT \cite{sun2021chinesebert} & \multicolumn{1}{c}{\textbf{57.03}} & \multicolumn{1}{c}{91.11} & \multicolumn{1}{c}{87.51} & \multicolumn{1}{c}{81.97} & 79.41 & 75.41  \\ \midrule
PHMOSpell \cite{huang2021phmospell} & \multicolumn{5}{c}{\multirow{3}{*}{/}} & 77.14 \\ \cmidrule{1-1} \cmidrule{7-7} 
ReadListenAndSee \cite{xu2021read} & \multicolumn{5}{c}{} & 77.84 \\ \cmidrule{1-1} \cmidrule{7-7} 
PLOME \cite{liu-etal-2021-plome} & \multicolumn{5}{c}{} & 77.20 \\ \midrule

PmBERT & \multicolumn{1}{c}{55.43} & \multicolumn{1}{c}{\textbf{91.88}} & \multicolumn{1}{c}{\textbf{88.23}} & \multicolumn{1}{c}{\textbf{82.95}} & \textbf{79.62} & \textbf{78.92} \\ 

\midrule\midrule
 \multicolumn{7}{c}{clean data}  \\ \cmidrule{1-6}
BERT \cite{devlin2018bert} & \multicolumn{1}{c}{64.48} & \multicolumn{1}{c}{95.14} & \multicolumn{1}{c}{95.17} & \multicolumn{1}{c}{89.99} & 86.20 & 1   \\ \cmidrule{1-6}
PERT$^*$ \cite{pbert} & \multicolumn{1}{c}{59.35} & \multicolumn{1}{c}{87.64} & \multicolumn{1}{c}{91.23} & \multicolumn{1}{c}{86.69} & 81.23 & 1 \\ \cmidrule{1-6}
BERT-from-PmBert & \multicolumn{1}{c}{64.45} & \multicolumn{1}{c}{94.28} & \multicolumn{1}{c}{94.69} & \multicolumn{1}{c}{88.44} & 85.47 & 1  \\ \cmidrule{1-6}
PmBERT & \multicolumn{1}{c}{64.77} & \multicolumn{1}{c}{94.94} & \multicolumn{1}{c}{94.57} & \multicolumn{1}{c}{89.60} & 85.97 & 1 \\ 
\bottomrule
\end{tabular}
}
\caption{Main result on NER, classification, and Chinese spell correction dataset, with F1 metric.}
\label{main_result}
\end{table*}
\begin{figure}
    \centering
    \includegraphics[width=7.7cm]{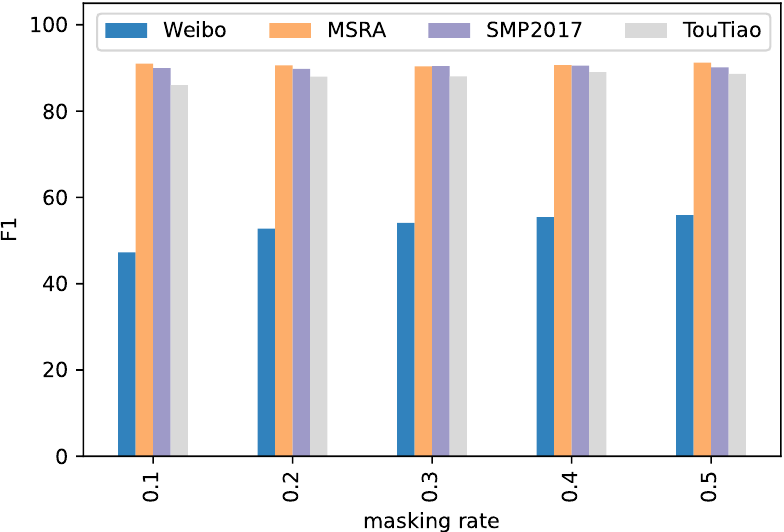}
    \caption{F1 results of different masking rate among 4 datasets.}
    \label{fig:masking_rate}
\end{figure}

\subsection{Main Experiments}
The results of our model are displayed in Table 3. From the table, the upper left half is the result of the noised dataset, the lower half is the result of the clean dataset, and the right is the result of the error correction dataset.

To evaluate the model's tolerance to SSP error, we firstly do experiments on the noisy NER and classfication dataset. As we can see from the upper left part of table, it beats all current methods, which verify the effectiveness of our model. In the finetuning stage, adding pinyin as an additional feature to BERT can enhance the SSP error tolerance, regardless by contacting or adding. While introducing pinyin in the pretraining stage can greatly improve the model's SSP error tolerance. Our method achieves a mean of 79.62 on four noisy datasets, an improvement of 11.08 percentage points over BERT's 68.74.

We also examined the error correction ability of the method on the Sighan15 dataset, which is the most commonly used Chinese error correction evaluation. Our results have exceeded SOTA models by 1.08 points, which further verifies its power.

Finally, we examine the model performance on the clean dataset. We can find that our model is slightly worse than BERT, which indicates that it keeps the full capabilities of BERT after adding pinyin. Even if only using characters without pinyin, our model's performance is still comparable to BERT. This characteristic is not present in the previous model.

\section{Discussion}
In this section, we do extensive experiments to explore what makes a robust phonetic enhanced Chinese language model. We adjust the masking strategy, representation of pinyin and pinyin fused ways to pretrain our model. Detailed experiments are presented in the following sections.

\begin{table}[]
\centering
\scalebox{0.88}{
\begin{tabular}{lcccc}
\toprule
Proportion & Weibo & MSRA & SMP2017 & Toutiao \\ \midrule
0.2-0.4-0.4 & 54.58 & 90.75 & 87.67 & 82.01 \\ \midrule
0.4-0.3-0.3 & 54.76 & 90.91 & 87.95 & 82.05 \\ \midrule
0.6-0.2-0.2 & 55.87 & 90.99 & 88.92 & 82.66 \\ \midrule
0.8-0.1-0.1 & 55.43 & 91.88 & 88.23 & 82.95 \\ 
\bottomrule
\end{tabular}
}
\caption{The result of different proportions of token and pinyin pretraining tasks. For example, 0.8-0.1-0.1 means 80\% time we mask token and pinyin together, 10\% time we mask token only, and 10\% time we mask pinyin only.}
\label{charPinyinRate}
\end{table}

\subsection{Masking strategy}
\paragraph{Masking rate} 
BERT suggests that 15\% is the best mask probability for masked language modeling pretraining. Later works \cite{sun2019ernie,liu2019roberta, joshi2020spanbert} following this settings. We wonder if this still works in the setting of our PmBERT. To investigate the effects of the mask probability, we pretrain our PmBERT with rate {10\%, 20\%, 30\%, 40\%, 50\%} and other hyperparameters keep identical. The results on noisy dataset(20\%noise) are showed in Figure \ref{fig:masking_rate}. As can be seen from the figure, on 4 datasets (two entity recognition, two text classification), the performance of the model is improved with the increase of the masking ratio. Therefore, when training with a pinyin mask, 15\% is still not enough to build a good relationship between characters and pinyin. Of course, it can't be too high. If the proportion of pinyin is too large, the ability of the model will be reduced. We suggest an appropriate mask prob should be between 40\% and 50\%.

\paragraph{Proportion of token and pinyin} 
As illustrated in section3.2, we have three kinds of masking tasks namely masking token and pinyin together, masking token only, and masking pinyin only. We want to explore whether the proportion of the three tasks influences the language model. We pretrain our PmBERT with the proportion {80\%, 10\%, 10\%}, {60\%, 20\%, 20\%}, {40\%, 30\%, 30\%}, and {20\%, 40\%, 40\%}, and the results on noisy dataset(20\%noise) are presented in Table  \ref{charPinyinRate}. It shows that the higher the proportion of words and pinyin masked together, the better the language model performs. As for the reason, we believe masking pinyin or words only will give the model a hint and make this task easier. The results also show that we still need a small portion of mask pinyin or token only task. Using pinyin to predict tokens and using words to predict pinyin actually bridge the relation between semantics and phonetics. We suggest to use {60\%-80\%, 10\%-20\%, 10\%-20\%} task proportion to train PmBERT.
\begin{table}[]
\scalebox{0.7}{
\begin{tabular}{lccccc}
\toprule
Model & \multicolumn{1}{c}{Weibo} & \multicolumn{1}{c}{MSRA} & \multicolumn{1}{c}{SMP2017} & \multicolumn{1}{c}{Toutiao} & Sighan \\ \midrule
 & \multicolumn{4}{c}{noisy data} &  \\ \midrule
PmBERT$_{rand}$ & \multicolumn{1}{c}{54.67} & \multicolumn{1}{c}{91.29} & \multicolumn{1}{c}{87.96} & \multicolumn{1}{c}{82.17} & - \\ \midrule
PmBERT$_{freq}$ & \multicolumn{1}{c}{55.43} & \multicolumn{1}{c}{91.88} & \multicolumn{1}{c}{88.23} & \multicolumn{1}{c}{82.95} & - \\ \midrule\midrule
 & \multicolumn{4}{c}{clean data} \\ \midrule
PmBERT$_{rand}$ & \multicolumn{1}{c}{65.11} & \multicolumn{1}{c}{94.29} & \multicolumn{1}{c}{94.24} & \multicolumn{1}{c}{88.68} & 77.68 \\ \midrule
PmBERT$_{freq}$ & \multicolumn{1}{c}{64.77} & \multicolumn{1}{c}{94.94} & \multicolumn{1}{c}{94.57} & \multicolumn{1}{c}{89.60} & 78.92 \\ 
\bottomrule
\end{tabular}
}
\caption{Results on different sampling strategies. Samling with frequency is slightly better.}
\label{freq}
\end{table}

\paragraph{Random sampling strategy} 

In the 10\% situation, compare to BERT randomly replacing token from vocab, we replace token from the phonetic candidate set according to their usage frequency. We want to know if the usage frequency benefits our model. We pretrain our PmBERT with the same configuration, and we randomly replace tokens from the phonetic candidate set. The results are represented in Table \ref{freq}. We can see freq sampling strategy is slightly better. These improvements may be due to the more adequate training of common words.

\subsection{Representation of pinyin}
Chinese pinyin generally consists of three parts: init, final, and tone. Take the Chinese character "中" for example, its pinyin can be represented as four kinds of format: zhong, zhōng, (zh, ong) and (zh, ong, 1), respectively pure pinyin, pinyin with tone, init and final, init and final with tone. In this section, we want to explore the impact of different representations of pinyin.

\paragraph{The impact of tone} 
For each Chinese pinyin, there would be four tones at most. To evaluate the effect of tone, we pretrain our PmBERT with a new pinyin vocab with tone. And all other hyperparameters remain the same. The results are presented in Table\ref{parts}. We can observe that pinyin without tones are superior to pinyin with tone across all tasks. This means tones are not very necessary when using pinyin in downstream tasks. In robustness settings, adding tone instead has the opposite effect.

\paragraph{The impact of init and final} 
When exploring the effects of init and final, we use three vocab embeddings to respectively represent init, final, and tone. The final pinyin representation would be the sum of the three . We pretrain our PmBERT under these settings, and the result is also shown in Table\ref{parts}. The results are very similar to the results of adding tones. The finer the pinyin segmentation, the worse the results.
The possible reason is that each pinyin can be more specifically mapped to the word by adding tones, which is the opposite of enhancing robustness. To enhance robustness, we hope that homophones have similar representations, which means that expressions should be more blurry. If it is divided into different parts, the dictionary expressing pinyin becomes trivial, and the relationship between pinyin and tokens cannot be well learned. This will weaken the ability of the language model to a certain extent.

\subsection{Mixed or parallel when pretraining}
There are three main ways to combine pinyin into language models in the pretraining stage. 
One is to indirectly use the confusion set as a candidate set to replace the original token when masking(1 in Figure\ref{mixed-or-parallel}). 
The second type is to directly replace the words of the original sentence with pinyin(2 and 3 in Figure\ref{mixed-or-parallel}); 
The third category is that words and pinyin are parallel, and participate in masking at the same time(4 in Figure\ref{mixed-or-parallel}).
Table\ref{pinyinUseWay} shows results on the noisy dataset with different using ways.

The mixed approach is better than the method that only uses a confusion set, and the parallel approach is the best.  The hybrid method cannot guarantee the consistency of inference and training which influence the performance in downstream tasks. Instead, the parallel method does not have this problem and has a better performance.
\begin{table}[]
\centering
\scalebox{0.8}{
\begin{tabular}{lcccc}
\toprule
Representaion & Weibo & MSRA & SMP2017 & TouTiao \\ \midrule
char & 49.50 & 71.61 & 80.57 & 73.28 \\ \midrule
\ \ +pinyin & 55.43 & 91.88 & 88.23 & 82.95 \\ \midrule
\ \ +pinyin+tone & 46.45 & 87.65 & 86.43 & 77.76 \\ \midrule
\ \ +init+final & 53.43 & 90.13 & 87.96 & 81.10 \\ \midrule
\ \ +init+final+tone & 45.63 & 85.92 & 86.37 & 80.58 \\ 
\bottomrule
\end{tabular}
}
\caption{The result of using different combinations of pinyin (init, final, tone) to pretrain. Char refers to the original BERT.}
\label{parts}
\end{table}
\begin{table}[]
\scalebox{0.88}{
\begin{tabular}{lcccc}
\toprule
Setting  & Weibo & MSRA & SMP2017 & TouTiao \\ \midrule
confusionset & 49.56 & 85.81 & 87.12 & 81.35 \\ \midrule
mixed & 49.82 & 84.99 & 87.81 & 80.74 \\ \midrule
mixed+out & 50.87 & 87.35 & 87.98 & 81.30 \\ \midrule
parallel & 54.13 & 90.02 & 87.64 & 81.37 \\ \midrule
parallel+out & 55.43 & 91.88 & 88.23 & 82.95 \\ 
\bottomrule
\end{tabular}
}
\caption{The result of pinyin fusing ways. The details of fusion way refer to Figure \ref{mixed-or-parallel} and Table \ref{use-way}.}
\label{pinyinUseWay}
\end{table}

\begin{table}[]
\centering
\scalebox{0.85}{
\begin{tabular}{lcccc}
\toprule
Data \textbackslash Nosie rate & \multicolumn{1}{c}{0\%} & \multicolumn{1}{c}{10\%} & \multicolumn{1}{c}{20\%} & 50\% \\ \midrule
 & \multicolumn{4}{c}{BERT} \\ \midrule
MSRA & \multicolumn{1}{c}{94.97} & \multicolumn{1}{c}{92.36} & \multicolumn{1}{c}{88.39} & 73.06 \\ \midrule
SMP2017 & \multicolumn{1}{c}{95.04} & \multicolumn{1}{c}{94.12} & \multicolumn{1}{c}{93.14} & 78.89 \\ \midrule\midrule
 & \multicolumn{4}{c}{PmBERT} \\ \midrule
MSRA & \multicolumn{1}{c}{95.12} & \multicolumn{1}{c}{92.22} & \multicolumn{1}{c}{93.35} & 90.64 \\ \midrule
SMP2017 & \multicolumn{1}{c}{94.68} & \multicolumn{1}{c}{94.58} & \multicolumn{1}{c}{93.39} & 90.65 \\ 
\bottomrule
\end{tabular}
}
\caption{With noise level increase, PmBERT show more and more robustnees to phonetic similar error than BERT.}
\end{table}

\subsection{Nosie rate}
Pinyin features are fused into our model for robustness to SSP problem. In this section, we want to explore what would PmBERT perform when we increase the noise level. We replace {0\%, 10\%, 20\%, 50\%} tokens of the downstream dataset to their confusion set to simulate noise level, finetune PmBERT with different noise level data, and the results are presented in Table 7. From the table, we notice that PmBERT is more and more stable than BERT with the level of noise increasing. It can resist the phonetic similar words attack.

\subsection{Embeddings visualization}

\begin{figure}[]
    \centering
    \includegraphics[width=0.45\textwidth]{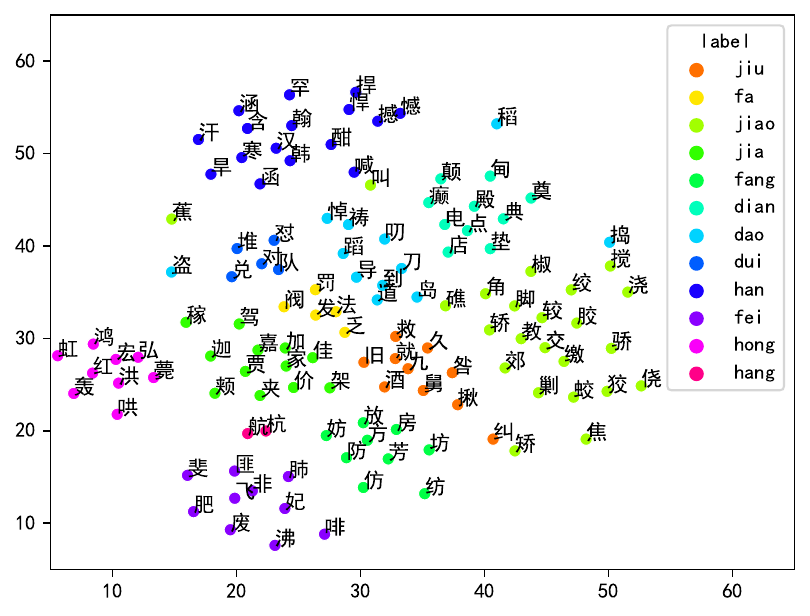}
    \caption{Characters embeddings and their corresponding pinyin. Characters with same color share same pinyin.}
    \label{fig:visual}
\end{figure}

The pretraining task is designed for bridging the relation between semantics and phonetics. We intuitively visualize the embeddings layer of our model to verify our proposed methods. Specifically, we use t-SNE \cite{van2008visualizing} to project the high dimensional token features to a 2d space. We select part of pinyin and their corresponding characters and the result is shown in Figure \ref{fig:visual}. We can find some phenomenons here 1) Chinese characters sharing the same phonetics tend to be close in the embeddings space, see the different color clusters. Take the orange dots in the figure for example, characters "就九旧酒舅揪咎救久" having the same pinyin "jiu" are very closed to each other. 2) The semantics in the original BERT model remains, e.g. character "喊"(shout) and character "叫"(shout) are very close in vector space despite they have different pinyin. These phenomena verify the effectiveness of our PmBERT. It has a strong power to model semantics and phonetics together.

\section{Related Work}
At present, there are relatively few studies on Chinese pinyin in pretraining language model. Existing related work mainly focuses on using pinyin to enhance the error correction ability of the model, or use pinyin to enhance the recognition ability in ASR systems. We can divide the work related to pinyin into two major categories, one is the indirect use of pinyin, and the other is the direct use, see table 1 for detail.
        
\paragraph{Indirect Use}
The indirect use of pinyin is mainly to add pinyin information to the model through data augmentation. Others use the candidate set of homophones to introduce pinyin information.
\citet{cui2021approach} propose a novel data augmentation method leveraging a pretraining language model to generate text containing ASR plausible errors. Their noisy data generator which is used to add noise to input text of downstream tasks are constructed through fine-tune GPT-2 \cite{radford2019language} on clean-noisy pairs. \citet{li2018improving} proposed four strategies to craft ASR-specific noise training data. They demonstrate that simple homophone-based substitution achieves the best performance.
\citet{wang2020data} use a data-driven approach rather than heuristic rules to simulate ASR errors. They collect pairs of ASR hypotheses and corresponding reference text, aligning them at the word level by minimizing the Levenshtein distance. Based on the aligned pairs, they count n-gram confusion frequencies which are used to replace n-grams in clean text using a confusion matrix to generate noisy text. \citet{simonnet-etal-2018-simulating}
generate ASR erroneous data through joint usage of both acoustic and linguistic word embeddings. They define a similarity metric that relates acoustic features and linguistic features to guide the generation of ASR error data. By training with the generated data, the performance of the SLU system is improved.
        
\paragraph{Direct Use} \citet{sun2021chinesebert} use a CNN model on pinyin sequence with the tone, contact pinyin features, and semantic features then use a fully connected layer to fuse features. Unlike conventional pretrain models, \citet{liu-etal-2021-plome,zhang2021correcting} use different mask schemes to bridge the relation between phonetic features and semantic features. \citet{liu-etal-2021-plome} pretrain language model with masks chosen from tokens with similar characters according to a confusion set, use GRU encoders to extract pinyin letters features and add pinyin features with character features. \citet{zhang2021correcting} mask characters with pinyin or similar pronounced characters to enable the language model to explore the similarities between characters and pinyin. 
        
\paragraph{Task specific model} \citet{huang2021phmospell} derive pinyin features from a TTS model Tacoton2\footnote{\url{https://github.com/NVIDIA/tacotron2}} and design a novel adaptive gating mechanism to fuse features. \citet{xu2021read} designs a hierarchical encoder to process pinyin letters at the character level and sentence level and uses a selective module to fuse features.

\section{Conclusion}
In this paper, we propose PmBERT, a pretraining method for inputting text and pinyin in parallel to make the Chinese language model robust to SSP problem. Through delicate designed pretraining tasks, our model can catch the relation between semantics and phonetics. In general, the performance of the pure pinyin model is not as good as that of the word model, but the word model with mixed pinyin can achieve considerable robustness to SSP problem. Through extensive experiments, we verify the effectiveness of our PmBERT. We also give detailed suggested hyperparameters to pretrain our model. These experiments on hyperparameters would be a backbone to further research in this area.

\section{Limitations}
Our model has shown its capability in robustness to SSP problem. But for the clean data, its performance does not make any improvement. More explorations should be done in the aspect of masking strategy, pinyin representation, and fusing methods.

\clearpage
\bibliographystyle{acl_natbib}
\bibliography{anthology,custom}

\clearpage
\appendix
\section{Masking Algorithm}
\label{sec:appendix}
\begin{algorithm}[t]
    \begin{normalsize}
        \SetAlgoNoLine  
        \caption{Masking Strategy}
        \label{alg:masking_strategy}
        get Masked candidate Masked\_candidate \;
        get $prob$ from a uniform distribution \;
        
        \For {$candidate \in Masked\_candidate$}
        {
            \eIf {$prob$  < mask\_prob}
            {
                \eIf  {$prob$  < mask\_prob * 0.8} {mask pinyin and token together}
                {
                    \eIf {$prob$  < mask\_prob * 0.9} {only mask token}
                    {only mask pinyin}
                }
                
            }
            {
                \eIf{ $prob$  < 0.9}
                {
                    sample a random token $t^\prime$ from confusion set by prob \;
                    replace token with $t^\prime$ \;
                }
                {unchanged}
            }
        }
  \end{normalsize}
\end{algorithm}

\section{Datasets}
\label{apd:datasets}
\paragraph{Weibo} 
is another Chinese NER dataset collected from a Chinese social media platform, Weibo. It contains 1890 messages. Annotation which include seven entity types follows the BIO standard. The dataset contains around 10k words. Roughly, 99.34\% of the weibo messages has length less than 128.

\paragraph{MSRA} 
is published for evaluating Chinses NER models. It is annotated in BIO format and contains around 50k sentences including four entity types: person, organization, location. 99.84\% sentences have length within size of 256.

\paragraph{SMP2017} 
is introduced for Chinese  human-computer dialogue systems. user intent classification of single utterance in. It contains 3,735 utterances with 31 kinds of intents. Only one utterance whose length is longer than 64.

\paragraph{Toutiao News} 
is provided by a Chinese News App, Toutiao. It contains around 38k Chinese short texts spanning 15 different domains. Only 8 sentences in the dataset have length longer than 128.

\paragraph{SIGHAN2015\cite{tseng2015introduction}}
This is a benchmark dataset for the evaluation of CGEC and it contains 2,339 samples for training and 1,100 samples for testing. As did in some typical previous works (\citet{wang2019confusionset}, \citet{li2021tail}), we also use the SIGHAN15 testset as the benchmark dataset to evaluate the performance of our models as well as the baseline methods in fixed-length (FixLen) error correction settings.

\end{CJK*}
\end{document}